\documentclass[conference]{IEEEtran}
\IEEEoverridecommandlockouts
\usepackage{cite}
\usepackage{amsmath,amssymb,amsfonts}
\usepackage{algorithmic}
\usepackage{graphicx}
\usepackage{textcomp}
\usepackage{xcolor}
\def\BibTeX{{\rm B\kern-.05em{\sc i\kern-.025em b}\kern-.08em
    T\kern-.1667em\lower.7ex\hbox{E}\kern-.125emX}}

\usepackage{comment}											
\usepackage{multirow}									        
\usepackage{adjustbox}											
\usepackage{makecell}											
\usepackage{algorithm}											
\usepackage{geometry}
\usepackage{subfigure}

\DeclareRobustCommand*{\IEEEauthorrefmark}[1]{%
	\raisebox{0pt}[0pt][0pt]{\textsuperscript{\footnotesize\ensuremath{#1}}}}

\begin{document}

\title{A Context Augmented Multi-Play Multi-Armed Bandit Algorithm for Fast Channel Allocation in Opportunistic Spectrum Access{\tiny}}

	\author{
		\IEEEauthorblockN{
			Ruiyu Li\IEEEauthorrefmark{1} \thanks{E-mails: ruiyli@stu.xidian.edu.cn (R. Li), gxli@xidian.edu.cn (G. Li), xiao.lu@ericsson.com (X. Lu), liujichao09@haier.com (J. Liu), jinyan11@haier.com (Y. Jin).},
			Guangxia Li\IEEEauthorrefmark{1\ *} \thanks{* The corresponding author.},
			Xiao Lu\IEEEauthorrefmark{2},
			Jichao Liu\IEEEauthorrefmark{3}
			and Yan Jin\IEEEauthorrefmark{3}}
		\IEEEauthorblockA{\IEEEauthorrefmark{1}School of Computer Science and Technology, Xidian University, China}
		\IEEEauthorblockA{\IEEEauthorrefmark{2}Research and Development, Ericsson, Canada}
		\IEEEauthorblockA{\IEEEauthorrefmark{3}Hainayun IoT Technology Ltd, China}
	}

\maketitle

\begin{abstract}
We study the restless contextual multi-play multi-armed bandit (MP-MAB) problem for channel allocation in the opportunity spectrum access (OSA) scenario.
Most existing MP-MAB methods are impractical for real-world OSA systems as they assume many ideal conditions, incur a heavy computational cost, and most importantly, ignore the impact of channel noise which is directly related to the quality of service.
In this study, we embody this impact by modeling channel noise as a perturbation of the arm's reward function in MP-MAB.
As there is an implicit correlation between channel state information and channel noise, we take the former as a context for MP-MAB to present the perturbation caused by the latter.
We investigate two types of correlation between the context and the perturbation---linear and nonlinear, and derive two index policies, respectively.
These policies learn the correlations through a linear model and a neural network, and use estimated noise value to adjust the upper confidence bound.
Numerical experiments demonstrate that the proposed policies can achieve lower regret and select sub-optimal arms in a more reasonable way.
\end{abstract}


\section{Introduction}
\label{sec1}

As a dynamic spectrum access technique, opportunistic spectrum access (OSA) allows secondary users (SUs) to opportunistically access spectrum bands of primary users (PUs) that are temporally idle, thereby improving spectrum utilization at a small cost.
It has been successfully applied in recent wireless communication networks like 5G mobile network~\cite{9161990}, UAV network~\cite{8918303}, and IoT network~\cite{aloqaily2020multi}.
Despite its appealing features and broad application prospects, the approach to allocating available spectrum resources to satisfy SUs' demands for transmission while ensuring sufficient protection for PUs' remains a challenging problem.

Consider an OSA scenario where a base station (BS) has to allocate $K$ channels to $M$ SUs for continuous data transmission. 
Such a sequential decision-making problem can be formulated as a multi-armed bandit (MAB) game in which a learner selects an arm (a channel) in each round by trading off between exploration and exploitation in order to maximize the cumulative reward for a long term. 
For an OSA system that involves multiple users (or put another way, that has to select more than one channel at each time), it evolves into what is known as the multi-play multi-armed bandit (MP-MAB), where the learner selects multiple arms instead of just one at each time step. 
The MP-MAB problem is more complex than the single-play version as the regret is not uniquely determined by the number of sub-optimal arm draws but also depends on the combinatorial structure of arm draws~\cite{pmlr-v37-komiyama15}.

When applied to OSA scenarios, most existing MP-MAB methods either depend too much on ideal preconditions or are computationally infeasible for real-world applications. 
For example, it has been required that each SU is aware of the states of others, thus imposing the system sensing and communication overheads, which may be unacceptable in practice~\cite{pmlr-v83-besson18a,boursier2019sic,pmlr-v48-rosenski16}. 
More importantly, all of the above methods assume that the reward distribution for each channel is stationary, while in a more realistic setting, the channel conditions may continue to change regardless of the learner's actions. 
Such a varying reward distribution can be modeled by a finite-state discrete-time Markov chain and processed by a particular kind of algorithm called the restless bandit.

For the restless MP-MAB, due to the continuously evolving nature of arm states (or, equivalently, the non-stationary distribution for generating channel payoffs), it is difficult to estimate the upper confidence bound (UCB) for each arm, which is essential for UCB-based methods.
A solution named regenerative cycle algorithm (RCA)~\cite{6200864} tries to construct small but fixed-state episodes using elaborately designed blocks that record arm rewards generated under the same state.
Because of the renewal property of Markov chain, the discrete blocks belonging to a single state can be concatenated into a consecutive sample path so that the mean reward of this path can be calculated as usual.
The RCA algorithm works effectively when the number of arm states is small (e.g., the experiment reported in~\cite{6200864} was conducted on a two-state Gilbert–Elliot channel model).
It suits the centralized OSA scenario where a BS allocates channels for SUs.

Previous studies have demonstrated that incorporating contextual feedback has a positive effect on making policies for bandit problems, especially for the restless case where the context information weakens the uncertainty of environment~\cite{10.1145/1870103.1870106,9578933}.
Inspired by this, we propose a context augmented restless MP-MAB algorithm to tackle the channel allocation problem in a time-varying OSA scenario.
As there are channel noises in the system and the noises disturb the channel payoff, we model the channel noise by means of the continuous distributed channel state information (CSI).
The algorithm takes CSI as context and blends it into the RCA framework so that the arm rewards are determined by the channel state and the context together.
The channels are allocated using two novel index policies, MP-LUCB and MP-NUCB, which can learn the correlation between context and perturbation through a linear model and a neural network, respectively, and the adjust UCB by considering channel noise.
Numerical experiments demonstrate that the proposed algorithms have lower regret compared to several context-free UCB methods.
Meanwhile, the sub-optimal arms it selects also exhibit a more reasonable structure.

The remainder of this paper is organized as follows.
Section~\ref{sec2} reviews related work.
Section~\ref{sec3} provides a problem formulation in terms of MP-MAB.
Section~\ref{sec4} presents two MP-MAB algorithms with linear and non-linear context models, respectively.
Section~\ref{sec5} gives experimental results and discussion.
Finally, Section~\ref{sec6} concludes this study.
\section{Related Work}
\label{sec2}

The restless MP-MAB has been applied in the OSA system where a BS allocates $K$ different channels to $M$ SUs for data transmission.
Each channel has either a \textit{good} or \textit{bad} state and can be represented as a restless bandit arm characterized by a two-state Markov chain.
It is desired to select sub-optimal arms (i.e., arms with small rewards) as few as possible to minimize the regret of MP-MAB.
To this end, several index policies have been proposed to calculate the UCB-like indices for each arm and select the top-$M$ arms~\cite{8868112,6200864,5394517}.
The restless nature implies that the reward distribution of arm may change even if it has not been selected.
It breaks the i.i.d. assumption on the reward process that ordinary bandit policies depend on.
To overcome this challenge, RCA~\cite{6200864} concatenates blocks with the same state for each arm to construct a consecutive sample path whose statistics are essentially the same as the actual path observed by the arm.
The sample mean rewards derived from the consecutive path can approximate the reward stationary distribution due to the renewal property of the Markov chain.

Channel noise has been seldom utilized by MAB-based channel allocation methods. 
A representative work partitions wireless channel noise on the basis of PU's transmission activity, small-scale fading and device mobility, and takes the latter as finite discrete contexts to assist a contextual MAB to allocate channels for SUs~\cite{9578933}.
Similarly, channel noise has been described in terms of channel fast fading, bursty interference, and other context information to optimize an allocation policy~\cite{8792155}.
For a realistic system whose environment is dynamically varying, one has to augment MP-MAB with contextual information in the restless setting~\cite{9781393,9721692}.
It has been attempted to use contexts such as user location and user-centric quality of experience to aid dynamic channel allocation for large-scale and high-mobility networks~\cite{8972418}.
In this study, we devise a continuous context augmented restless MP-MAB to guide a BS to allocate channels for SUs in the OSA system.
\section{Preliminaries}
\label{sec3}

In an OSA scenario, a BS allocates $M$ channels for SUs from $K$ candidate channels with spectrums $(K \geq M)$ for data transmission at each time step.
We model this problem as a restless multi-play multi-armed contextual bandit.
Formally, we assume that there are $K$ channels with ideal spectrum, corresponding to $K$ arms in the contextual bandit indexed by $k \in \left\{1, \cdots, K \right\}$.
Each arm $k$ is characterized by $\left(n_k, MC_k\right)$, where $n_k$ represents the noise that the channel $k$ suffers, and $MC_k$ is a discrete-time, irreducible, and aperiodic Markov chain with a finite state space $S_k$.
The stationary distribution of the $MC_k$ is denoted as $\alpha_k$.
The mean reward $\mu_k$ of playing arm $k$ is defined as:
\begin{equation}
	\label{eq:meanreward}
	\mu_k = \sum_{s \in S_k} h_k (s) \alpha_k(s) - n_k
\end{equation}
where $h_k(s)$ denotes the reward of playing arm $k$ when its state is $s$.
It is assumed that all $\mu_k$ for $k \in \left\{1,\cdots,K \right\}$ are distinct and arranged in descending order as $\mu_1 \geq \mu_2 \geq \cdots \geq \mu_K$.
In this way, $\left\{ 1, 2, \cdots, M \right\}$ is the set of $M$ optimal arms, and $\left\{ M+1, M+2, \cdots, K \right\}$ is the set of sub-optimal arms.

Consider an episodic task of $T$ time steps.
At each time step $t \left(1 \leq t \leq T \right)$, the learner (BS) selects $M$ arms from $K$ candidates simultaneously without the knowledge of neither arms' mean rewards nor their ordered sequence.
Let $\mathcal{A}^t = \left\{a_1^t, a_2^t, \cdots, a_M^t \right\}$ denote the set of selected arms at time step $t$, where $a_m^t \in \left\{1, \cdots, K \right\}$ and $m = 1, 2, \cdots, M$.
After making selections, the learner receives rewards $r \left( h_{a_m}(s^t), n_{a_m}^t \right)$, which are independent across arms.
To evaluate the learner's policy, we define regret as:
\begin{equation}
	\label{eq:regret}
	R(T) = T \sum_{m=1}^{M} \mu_m - \mathbb{E} \left[ \sum_{t=1}^{T} \sum_{a \in \mathcal{A}^t} \left( h (s_{a}^t) - n_{a}^t \right) \right]
\end{equation}
It measures the gap between the best total reward (gained by playing $M$-optimal arms only ) and the accumulated total reward.
\section{Contextual MP-MAB}
\label{sec4}

We study the channel allocation policy under the effect of channel noise.
As aforementioned, such a problem can be modeled as a contextual MP-MAB where at each time step, the learner observes $K$ context vectors for each of the $K$ arms and selects $M$ ($M \leq K$) arms by applying the policy.
The context vector reflects the perturbation of arm's reward which is the effect of channel noise.
We derive two policies---MP-LUCB and MP-NUCB based on the assumption that the correlation between context and perturbation is linear and nonlinear, respectively.
The following begins with an elaboration of MP-LUCB.
Subsequently, a brief introduction to MP-NUCB is presented.

\subsection{MP-LUCB}

At each time step $t$, the learner observes $K$ context vectors $\mathbf{x}_k^t \in \mathbb{R}^d$ to characterize the arm perturbations where $\Vert \mathbf{x}_k^t \Vert \leq 1$, and $\Vert \cdot \Vert$ denotes the $\ell_2$-norm.
Subsequently, it selects $M$ arms according to a custom upper confidence bound that is aware of perturbation~\cite{pmlr-v15-chu11a}, and then receives the reward corresponding to its selection.
The assumed linear correlation between reward perturbations and context vectors is parameterized by $\mathbf{\theta}_* \in \mathbb{R}^d$ with $\Vert \mathbf{\theta}_* \Vert \leq 1$.
Following~\cite{pmlr-v15-chu11a}, the expectation of the perturbation of the $k$-th arm reward is
\begin{equation}
	\label{eq:noise}
	\mathbb{E} \left[ n_k^t \mid \mathbf{x}_k^t \right] = \mathbf{\theta}_*^{\top} \mathbf{x}_k^t
	\notag 
\end{equation}
By this way, we can rewrite Eq.~\eqref{eq:regret} as:
\begin{equation}
	\label{eq:reregret}
	R(T) = T \sum_{m=1}^{M} \mu_m - \mathbb{E} \left[ \sum_{t=1}^{T} \sum_{a \in \mathcal{A}^t} \left( h (s_{a}^t) - \mathbf{\theta}_a^{\top} \mathbf{x}_a^t \right) \right]
	\notag 
\end{equation}

Denote the context matrix for all arms by $\mathbf{X}^t = \left\{ \mathbf{x}_1^t, \mathbf{x}_2^t, \cdots, \mathbf{x}_K^t \right\} \in \mathbb{R}^{K \times d}$, and the reward vector for all arms by $\mathbf{r}^t = \left\{ r_1^t, r_2^t, \cdots, r_K^t \right\} \in \mathbb{R}^K$, where $r_i = h(s_i) - \mathbf{\theta}_i^{\top} \mathbf{x}_i$.
We can use Ridge Regression~\cite{10.5555/1795114.1795183} to approximate the unknown parameter $\mathbf{\theta}_*$.
The corresponding loss function can be defined as:
\begin{equation}
	\mathcal{L} (\mathbf{\theta}) = \Vert \mathbf{X} \mathbf{\theta} - \mathbf{r} \Vert^2 + \Vert \mathbf{I} \mathbf{\theta} \Vert
	\label{op:rlf}
\end{equation}

Let $\mathbf{A} = \left( \mathbf{X}^{\top}\mathbf{X} + \mathbf{I}^{\top}\mathbf{I} \right)^{-1}$ and $\mathbf{b} = \mathbf{X}^{\top} \mathbf{r}$, the solution of $\mathbf{\theta}$ in Eq.~\eqref{op:rlf} is $\mathbf{A}\mathbf{b}$.
Since there is a bias between the estimated perturbation obtained by the linear model ($\mathbf{\theta}_k^{\top} \mathbf{x}_k^t$) and the actual value, we calibrate the estimation with confidence intervals as suggested in~\cite{pmlr-v119-zhou20a}.

\begin{algorithm}[htpb]
	\caption{Multi-Play Linear UCB (MP-LUCB)}
	\label{alg:algorithm1}
	\begin{algorithmic}[1]
		\STATE \textbf{Inputs:} context vector dimension $d$, channel number $K$, SU number $M$, hyperparameter $\beta$\
		\STATE \textbf{Initialize:} $\mathbf{A} = \mathbf{I} \in \mathbb{R}^{d \times d}, \mathbf{b} = \mathbf{0} \in \mathbb{R}^d$\
		
		\FOR{$t = 1, 2, \cdots, T$}
		\STATE Observe the channel context vectors $\mathbf{x}_k^t \in \mathbb{R}^d$\
		
		\FOR{$k = 1, 2, \cdots, K$}		
		\STATE Observe the channel state $s_k^t$ according to the Markov chain evolution\
		\STATE Compute the parameters of the linear model $\mathbf{\theta}_k = \mathbf{A}_k \mathbf{b}_k$\
		\STATE Compute the perturbation-aware upper confidence bound $u_k^t$ according to Eq.~\eqref{eq:linucb}\
		\ENDFOR
		
		\STATE Sort $K$ channels in descending order of $u_k^t$\
		\STATE Select the first $M$ channels as candidate set $\mathcal{A}^t$, and the rest as complement set $\mathcal{C}^t$\
		
		\FOR{$k = 1, 2, \cdots, K$}
		\STATE Compute the channel reward $r_k = h(s_k) - n(\mathbf{\theta}_k)$\
		\FOR{$m \in \mathcal{A}^t$}
		\IF{$r_m \leq 0$ and $j \in \mathcal{C}^t$ and $r_j \geq 0$}
		\STATE Replace $m$ with $j$\
		\ENDIF
		\ENDFOR
		\ENDFOR
		
		\FOR{$m \in \mathcal{A}^t$}
		\STATE Update $\mathbf{A}_m = \mathbf{A}_m + \mathbf{x}_m^t \left( \mathbf{x}_m^t \right)^{\top}$\
		\STATE Update $\mathbf{b}_m = \mathbf{b}_m + \left(\mathbf{\theta}_k^{\top} \mathbf{x}_m^t \right) \mathbf{x}_m^t$\
		\ENDFOR
		\ENDFOR
	\end{algorithmic}
\end{algorithm}

Assume that the value $\hat{n}$ obtained from the linear regression deviates from the actual perturbation $n$ by a value of $\delta$.
That is, $\hat{n} - \delta < n < \hat{n} + \delta$.
As Eq.~\eqref{eq:meanreward} has suggested, the basis for selecting arms is the difference of reward and the perturbation $n$.
We only consider an overestimation of $n$, i.e., $n = \hat{n} + \delta$.
This is because if an arm can be selected in such a maximum perturbation case, it will be selected when the perturbation is mild.
Referring to~\cite{10.5555/1795114.1795183}, we can derive the solution for $\delta$ as:
\begin{equation}
	\delta = \beta \sqrt{\left( {\mathbf{x}_k} \right)^{\top} \mathbf{A}^{-1} \left( {\mathbf{x}_k} \right)}
	\notag 
\end{equation}
where $\beta$ is a hyperparameter.

We use the derived $\delta$ to calibrate the estimated perturbation value, and calculate the upper confidence bound (UCB) in a way that is proposed in~\cite{10.5555/1795114.1795183}, as follows:
\begin{equation}
	\label{eq:linucb}
	u_k = \mathbf{\theta}_k^{\top} \mathbf{x}_k + \beta \sqrt{\left( {\mathbf{x}_k} \right)^{\top} \left( {\mathbf{A}}\right)^{-1} \left( {\mathbf{x}_k} \right)}
\end{equation}

The proposed Multi-Play Linear UCB (MP-LUCB) is summarized in Algorithm~\ref{alg:algorithm1}.
It iteratively receives context from outside, and uses the obtained context to update a linear model to estimate perturbation of channel reward.
The estimated perturbation value, after bias correction, is used to calculate the confidence upper bound (\textit{Line 8}).
The top-$M$ channels with high UCB are then selected as candidates for $M$ SUs (\textit{Line 11}).
As the channel reward calculated as in \textit{Line 13} may be negative, we make a remedy by replacing a candidate member with a complement one (\textit{Line 14-18}).
The resulting candidates are assigned to SUs as the allocated channels.

\subsection{MP-NUCB}

The above MP-LUCB demonstrates the idea of learning the correlation between context and perturbation, but its linear context model may be over simplified.
In a real-world OSA system, there are complex correlations between channel noise and channel state information.
An influential work~\cite{8640815} employs neural network to model the nonlinear correlation between the time-frequency response of a fast-fading communication channel and its quality of service, and demonstrates that the resulting estimation leads to better channel allocation.
Inspired by such, we use neural network to embed the context vector into a latent space to explore the intrinsic correlation of context and perturbation.
The output perturbation value is used to construct the upper confidence bound in a similar way as Eq.~\eqref{eq:linucb}.

Formally, for a neural network of $L$ layers and width $D$, the perturbation $n_k$ is learned as follows:
\begin{equation}
	n_k \left( \mathbf{x}_k; \mathbf{\theta}_k \right) = \mathbf{W}_L \sigma \left( \mathbf{W}_{L-1} \sigma \left( \cdots \sigma \left( \mathbf{W}_1 \mathbf{x}_k \right) \right) \right)
	\notag 
\end{equation}
where $\sigma$ is the rectified linear unit (ReLU) activation function, $\sigma(z) = \max \left\{ z, 0 \right\}$, $\mathbf{W}_L \in \mathbb{R}^{D \times 1}$, $\mathbf{W}_1 \in \mathbb{R}^{D \times d}$, and $\mathbf{W}_i \in \mathbb{R}^{D \times D}$ for $2 \leq i \leq L-1$.

Similar to Eq.~\eqref{eq:linucb}, the upper confidence bound can be derived as:
\begin{equation}
	u_k = n_k \left( \mathbf{x}_k; \mathbf{\theta}_k \right) + \gamma \sqrt{\left( \mathbf{g}_k \right)^{\top} \left( {\mathbf{A}}\right)^{-1} \left( \mathbf{g}_k \right)}
	\notag 
\end{equation}
where $\mathbf{g}_k$ is used to perform gradient descent for updating $\mathbf{\theta}_k$, and $\mathbf{g}_k = \mathbf{g} \left( \mathbf{x}_k; \mathbf{\theta}_k \right) = \nabla_{\mathbf{\theta}_k} n_k \left( \mathbf{x}_k; \mathbf{\theta}_k \right)$.

The matrix $\mathbf{A}$ is then updated using $\mathbf{g}_k$ as follows:
\begin{equation}
	\mathbf{A} = \mathbf{A} + \mathbf{g} \left( \mathbf{x}_k; \mathbf{\theta}_k \right) \mathbf{g} \left( \mathbf{x}_k; \mathbf{\theta}_k \right)^{\top} / D
	\notag 
\end{equation}

Regarding the parameter's initialization, we follow \cite{pmlr-v119-zhou20a} to initialize the network by generating each entry from a Gaussian distribution.
Specifically, we set
\begin{equation}
	\mathbf{W}_l = \begin{bmatrix}
		\mathbf{W} & \mathbf{0} \\ 
		\mathbf{0} & \mathbf{W} 
	\end{bmatrix}
	\notag 
\end{equation}
for $1 \leq l \leq L-1$, where each entry of $\mathbf{W}$ is generated independently from $\mathcal{N} (0, 4/D)$; and $\mathbf{W}_L = \left[ \mathbf{w}^{\top}, -\mathbf{w}^{\top} \right]$, where each entry of $\mathbf{w}$ is generated independently from $\mathcal{N} (0, 2/D)$.
\section{Experiment Results}
\label{sec5}

We perform a numerical study by simulating four OSA scenarios with different state transition probabilities as in~\cite{6200864}.
Each scenario has ten Gilbert-Elliot channels, i.e., channels with two states, \textit{good} and \textit{bad}, which are denoted as $1$ and $0$, respectively.
The reward of channel $k$ is determined by its state and the noise, i.e., if $\mathbb{I} (s_k=1) - n_k > 0$, $r_k = 1$; otherwise $r_k = 0.1$, where $\mathbb{I}$ is an indicator function.
Following~\cite{6200864},  we denote the four scenarios as $S1$, $S2$, $S3$, and $S4$.
The state transition probabilities and the mean rewards of channels in each scenario are listed in Tables~\ref{tab1} and~\ref{tab2}, respectively.
The proposed MP-LUCB and MP-NUCB are compared with the following baselines:
\begin{itemize}
	\item MP-Random: Channels assigned to SUs are selected randomly;
	\item MP-UCB: A multi-play version of the canonical UCB where the top-$M$ arms are selected~\cite{pmlr-v28-chen13a};
	\item MP-KLUCB: A multi-play bandit using KL-divergence amended UCB as its policy~\cite{pmlr-v37-komiyama15};
	\item RCA-M: A multi-play version of the regenerative cycle algorithm that is dedicated to solve the restless bandit problem~\cite{6200864}.
\end{itemize}

\subsection{Parameter Settings}

The OSA scenario we consider includes ten channels ($K=10$) and five SUs ($M=5$).
The context is represented as an eight-dimension vector ($d=8$) whose elements satisfy a uniform distribution $U(0, 1)$.
For MP-LUCB, its hyperparameter $\beta$ is selected from a grid search over $\left\{ 0.1, 0.5, 1.0, 10, 100, 1000 \right\}$.
For MP-NUCB, the underlying neural network's layers and width are set as $L = 2$ and $D = 16$, respectively.
And a dropout with $p = 0.1$ is applied between these two layers.
The network is optimized using the Adam with a learning rate of $0.005$.

The simulation has been repeated five times and each time there are $T = 10^5$ time steps.
All results reported below are averaged over five repetitions.
To cope with the restless nature of simulated environment, we choose to evaluate MP-LUCB and MP-NUCB under the framework of RCA.
Therefore, both MP-LUCB and MP-NUCB share the same settings with RCA-M as has stated in~\cite{6200864}.

\begin{table}[t]
	\caption{Transition Probabilities of Channels in Four Scenarios~\cite{6200864}}
	\begin{center}
		\setcellgapes{0.4ex}
		\makegapedcells
		\begin{adjustbox}{width=\textheight / 3}
			\begin{tabular}{c|c|c|c|c|c|c|c|c|c|c|c}
				\hline
				\multicolumn{2}{c|}{\text{Channel}} & $1$ & $2$ & $3$ & $4$ & $5$ & $6$ & $7$ & $8$ & $9$ & $10$ \\
				\hline
				\multirow{2}{*}{$S1$} & $p_{01}$ & $0.01$ & $0.01$ & $0.02$ & $0.02$ & $0.03$ & $0.03$ & $0.04$ & $0.04$ & $0.05$ & $0.05$ \\
				& $p_{10}$ & $0.08$ & $0.07$ & $0.08$ & $0.07$ & $0.08$ & $0.07$ & $0.02$ & $0.01$ & $0.02$ & $0.01$ \\
				\hline
				\multirow{2}{*}{$S2$} & $p_{01}$ & $0.1$ & $0.1$ & $0.2$ & $0.3$ & $0.4$ & $0.5$ & $0.6$ & $0.7$ & $0.8$ & $0.9$ \\
				& $p_{10}$ & $0.9$ & $0.9$ & $0.8$ & $0.7$ & $0.6$ & $0.5$ & $0.4$ & $0.3$ & $0.2$ & $0.1$ \\
				\hline
				\multirow{2}{*}{$S3$} & $p_{01}$ & $0.01$ & $0.1$ & $0.02$ & $0.3$ & $0.04$ & $0.5$ & $0.06$ & $0.7$ & $0.08$ & $0.9$ \\
				& $p_{10}$ & $0.09$ & $0.9$ & $0.08$ & $0.7$ & $0.06$ & $0.5$ & $0.04$ & $0.3$ & $0.02$ & $0.1$ \\
				\hline
				\multirow{2}{*}{$S4$} & $p_{01}$ & $0.02$ & $0.04$ & $0.04$ & $0.5$ & $0.06$ & $0.05$ & $0.7$ & $0.8$ & $0.9$ & $0.9$ \\
				& $p_{10}$ & $0.03$ & $0.03$ & $0.04$ & $0.4$ & $0.05$ & $0.06$ & $0.6$ & $0.7$ & $0.8$ & $0.9$ \\
				\hline
			\end{tabular}
		\end{adjustbox}
		\label{tab1}
	\end{center}
\end{table}

\begin{table}[t]
	\caption{Mean Rewards of Channels in Four Scenarios~\cite{6200864}}
	\begin{center}
		\setcellgapes{0.5ex}
		\makegapedcells
		\begin{adjustbox}{width=\textheight / 3}
			\begin{tabular}{c|c|c|c|c|c|c|c|c|c|c}
				\hline
				\text{Channel} & $1$ & $2$ & $3$ & $4$ & $5$ & $6$ & $7$ & $8$ & $9$ & $10$ \\
				\hline
				$S1$ & $0.2$ & $0.21$ & $0.28$ & $0.3$ & $0.35$ & $0.37$ & $0.7$ & $0.82$ & $0.74$ & $0.85$ \\
				\hline
				$S2$ & $0.19$ & $0.19$ & $0.28$ & $0.37$ & $0.46$ & $0.55$ & $0.64$ & $0.73$ & $0.82$ & $0.91$ \\
				\hline
				$S3$ & $0.19$ & $0.19$ & $0.28$ & $0.37$ & $0.46$ & $0.55$ & $0.64$ & $0.73$ & $0.82$ & $0.91$ \\
				\hline
				$S4$ & $0.46$ & $0.614$ & $0.55$ & $0.6$ & $0.591$ & $0.509$ & $0.585$ & $0.58$ & $0.577$ & $0.55$ \\
				\hline
			\end{tabular}
		\end{adjustbox}
		\label{tab2}
	\end{center}
\end{table}

\begin{figure*}[htbp]
	\centering
	\subfigure[$S1$]{
		\includegraphics[width=0.23\linewidth]{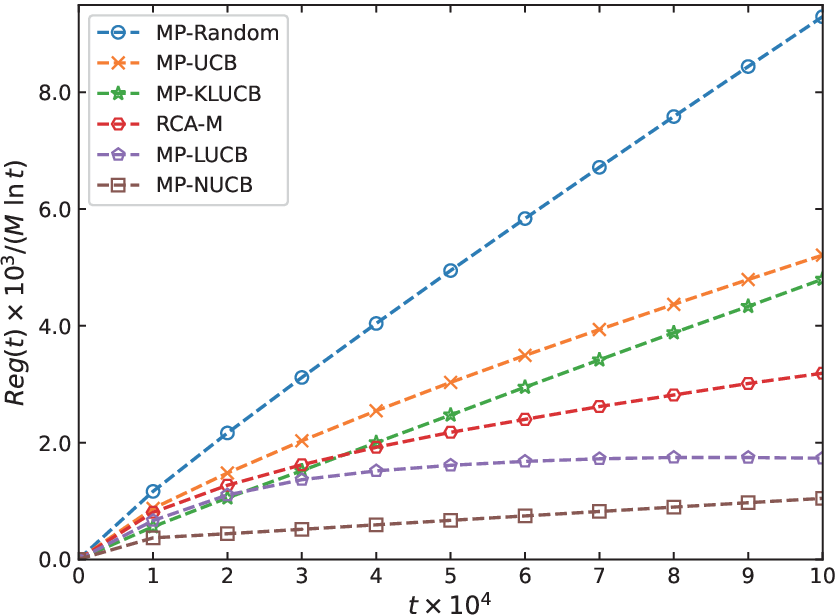}
		\label{fig1a}
	}
	\subfigure[$S2$]{
		\includegraphics[width=0.23\linewidth]{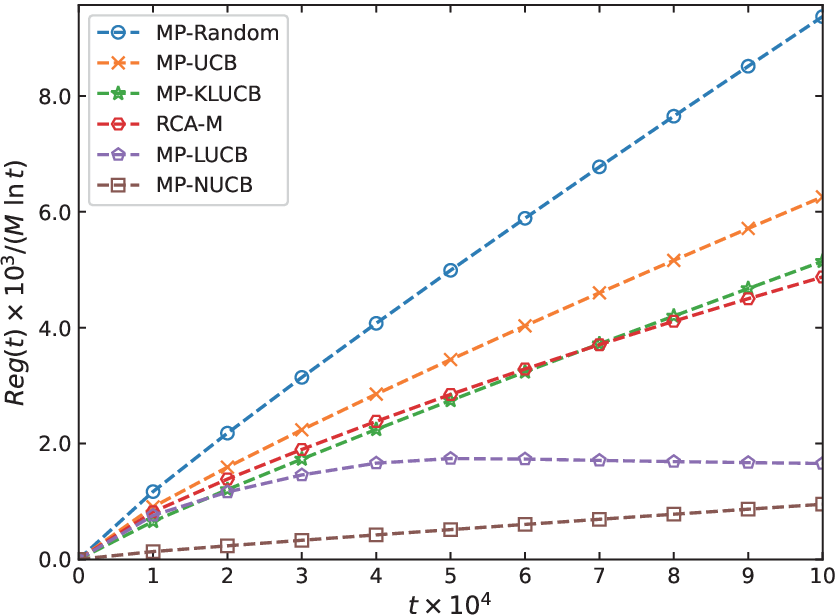}
		\label{fig1b}
	}
	\subfigure[$S3$]{
		\includegraphics[width=0.23\linewidth]{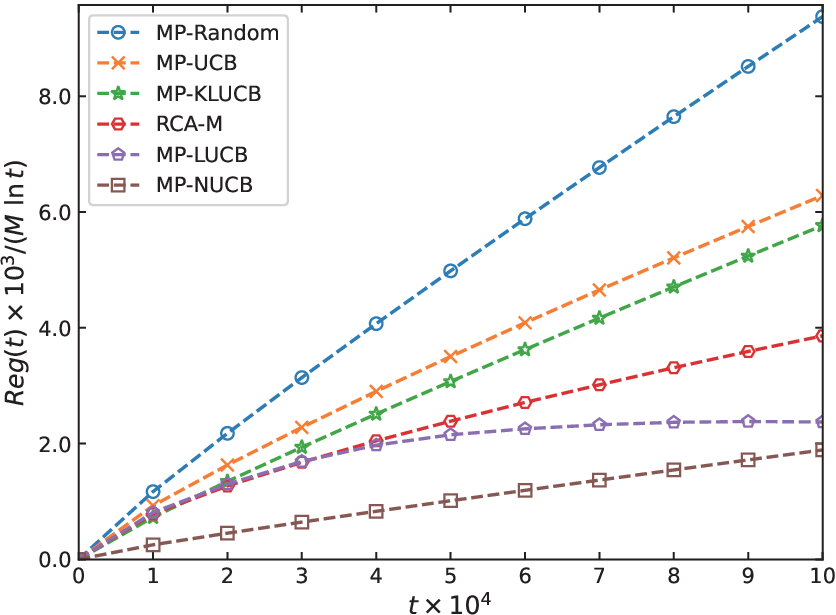}
		\label{fig1c}
	}
	\subfigure[$S4$]{
		\includegraphics[width=0.23\linewidth]{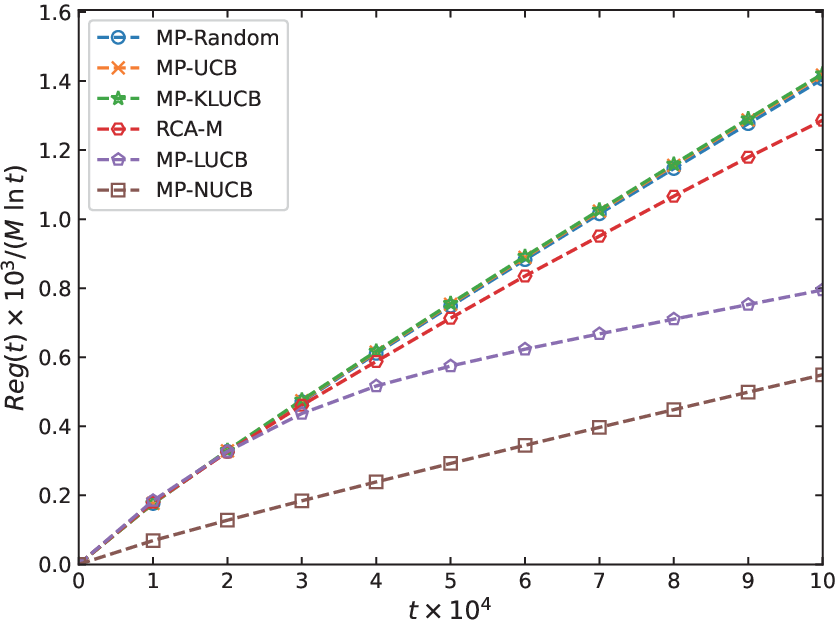}
		\label{fig1d}
	}
	\caption{Averaged variations of the regret, as defined in Eq.~\eqref{eq:regret}, over the number of selected channels ($M = 5$). The numerical results are normalized by $\ln t$ to make the curves more separable.}
	\label{fig1}
\end{figure*}

\begin{figure*}[htbp]
	\centering
	\subfigure[$S1$]{
		\includegraphics[width=0.23\linewidth]{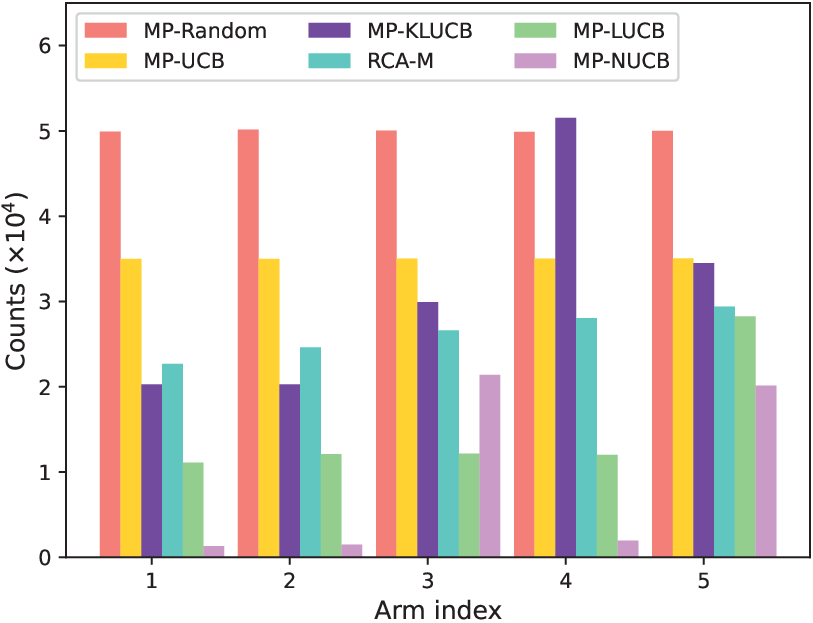}
		\label{fig2a}
	}
	\subfigure[$S2$]{
		\includegraphics[width=0.23\linewidth]{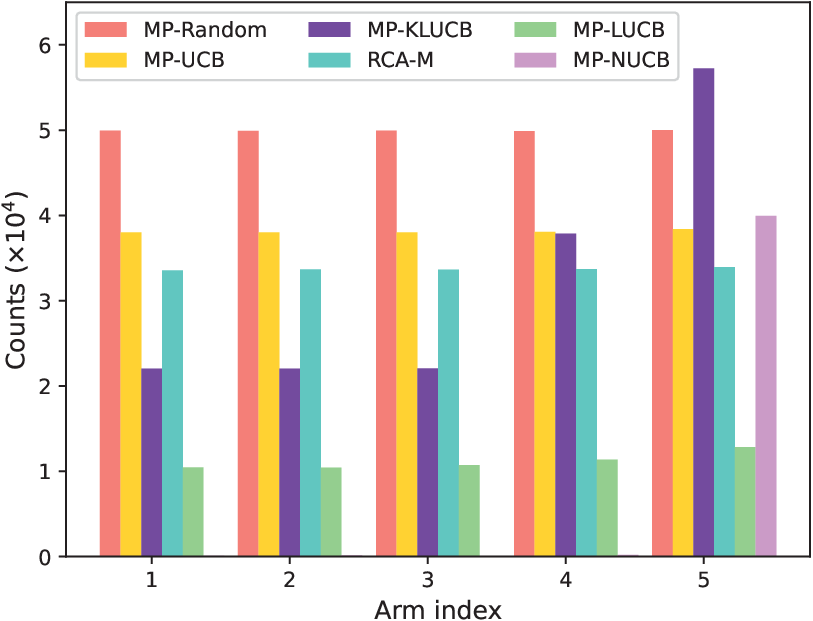}
		\label{fig2b}
	}
	\subfigure[$S3$]{
		\includegraphics[width=0.23\linewidth]{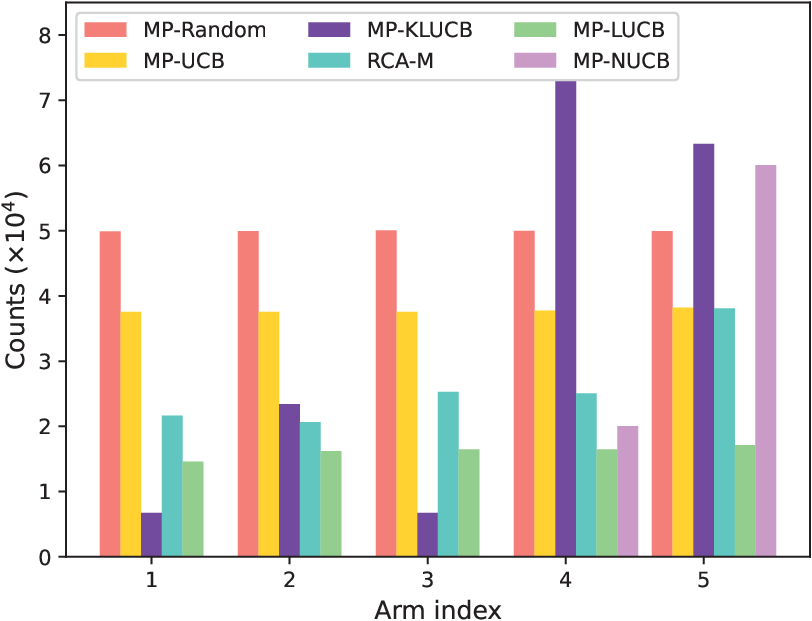}
		\label{fig2c}
	}
	\subfigure[$S4$]{
		\includegraphics[width=0.23\linewidth]{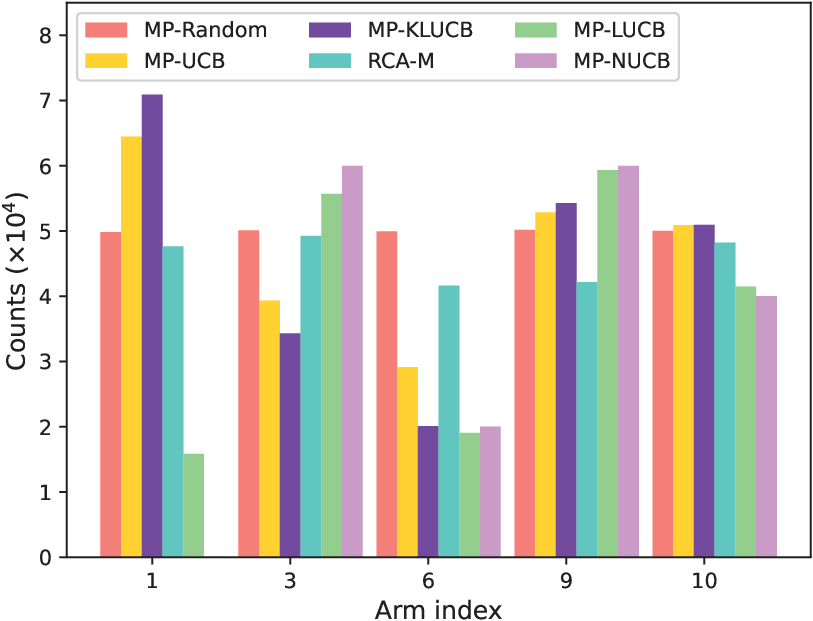}
		\label{fig2d}
	}
	\caption{Averaged number of times that sub-optimal channels are selected in $10^5$ time steps.}		
	\label{fig2}
\end{figure*}

\begin{figure}[htbp]
	\centering
	\subfigure[MP-LUCB]{
		\includegraphics[width=0.46\linewidth]{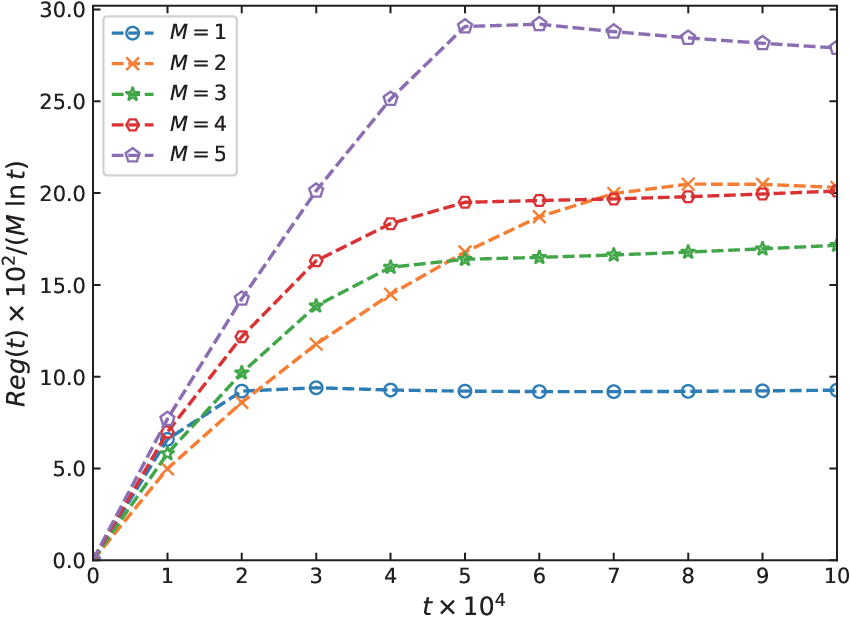}
		\label{fig3a}
	}
	\subfigure[MP-NUCB]{
		\includegraphics[width=0.46\linewidth]{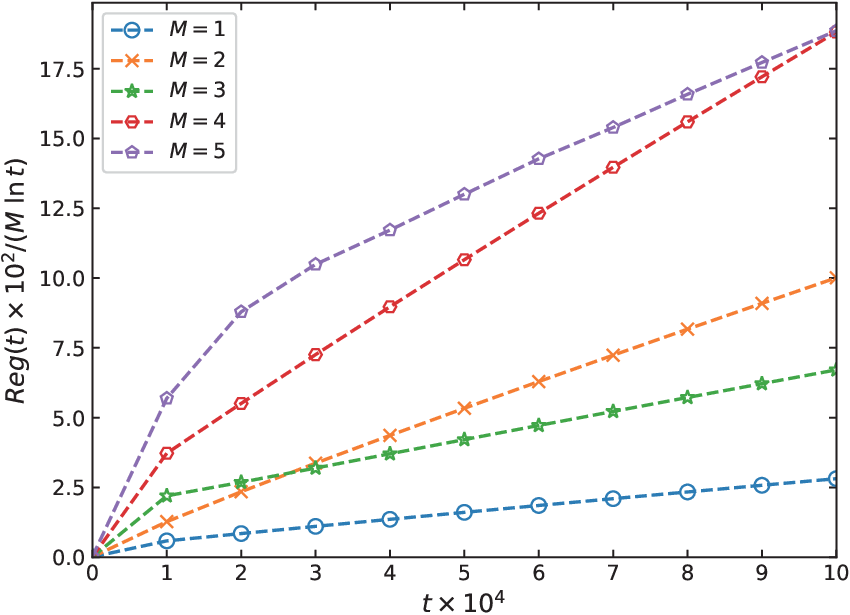}
		\label{fig3b}
	}
	\caption{The trend of normalized regret with different selected channel numbers $M$ in scenario $S1$.}
	\label{fig3}
\end{figure}

\subsection{Comparison of Regret}

Fig.~\ref{fig1} depicts the variation of the regret defined in Eq.~\eqref{eq:regret} averaged over the number of selected channels ($M = 5$).
It can be observed that the proposed MP-LUCB and MP-NUCB achieve the lowest regret, demonstrating that it is effective for the learner to incorporate context into policies.
To further examine the influence of selected channel number, we plot the trend of regret with different $M$ in Fig.~\ref{fig3}.
As is expected, allocating more channels increases the difficulty as the regret rises at the same time.

\subsection{Structure of Selected Arms}

As has stated in~\cite{pmlr-v37-komiyama15}, both the sub-optimal arm's selection times and the structure of sub-optimal arms affect the cumulative regret.
For illustration, consider scenario $S4$, whose channel rewards are listed in Table~\ref{tab2}.
In the case of $M=5$, a simple sort of reward values shows that the sub-optimal arms are $\left\{ 1,3,6,9,10 \right\}$.
Suppose there are two different policies: $policy_1$ and $policy_2$.
For $policy_1$, its selected arms in two successive time steps and the corresponding regrets are as follows:~\footnote{Take $Regret_{11}$ as an instance, the corresponding optimal channel combination is $\left\{ 2,4,5,7,8 \right\}$, and the actual selection is $\left\{ 1, 2, 3, 4, 5 \right\}$. Adding up the corresponding mean rewards according to Table~\ref{tab2}, and then subtracting gives the regret result.}
\begin{equation}
	\begin{aligned}
			selection_{11} &= \left\{ \underline{1}, 2, \underline{3}, 4, 5 \right\} &\Rightarrow Regret_{11} = 0.155 \\ 
			selection_{12} &= \left\{ \underline{6}, 7, 8, \underline{9}, \underline{10} \right\} &\Rightarrow Regret_{12} = 0.169
	\end{aligned}
	\notag
\end{equation}
For $policy_2$, the selections and regrets are
\begin{equation}
	\begin{aligned}
		selection_{21} &= \left\{ \underline{1}, 2, \underline{3}, \underline{6}, \underline{9} \right\} &\Rightarrow Regret_{21} = 0.260 \\
		selection_{22} &= \left\{ 2, 4, 5, 8, \underline{10} \right\} &\Rightarrow Regret_{22} = 0.035
	\end{aligned}
	\notag
\end{equation}

It can be seen that $policy_1$ and $policy_2$ select the same number of sub-optimal arms in total, but with different structures.
It is the different structures that lead to the difference in total regret of two policies ($0.324$ vs. $0.295$).

Referring to the proposed MP-LUCB and MP-NUCB, their regret curves in Fig.~\ref{fig1} indicate that they can select sub-optimal arms in a more reasonable way.
Besides, they also select fewer sub-optimal arms (see Fig.~\ref{fig2}). 
Both results demonstrate the advantage of our methods compared to baselines.

\subsection{Effect of Hyperparameter $\beta$}

\begin{figure}
	\centerline{\includegraphics[width=0.46\linewidth]{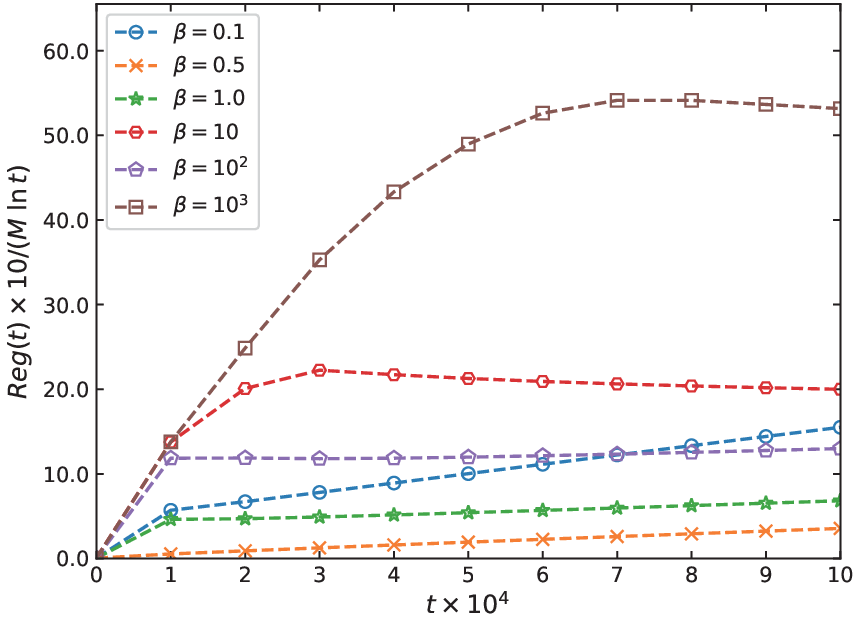}}
	\caption{MP-LUCB's normalized regrets under different settings of $\beta$ in scenario $S1$.}
	\label{fig4}
\end{figure}

The hyperparameter $\beta$ in Eq.~\eqref{eq:linucb} determines the confidence level of each channel, which consequently controls the degree of exploration by the learner.
A small $\beta$ value leads to a conservative policy that prefers to exploit channels with great estimated rewards, while a large value advocates exploring more channels.
Fig.~\ref{fig4} depicts the comparison of regrets under different settings of $\beta$.
It can be seen that neither a large $\beta$ nor a small one achieves a satisfying result.
As there is no prior knowledge, the hyperparameter $\beta$ should be tuned in practice. 

\section{Conclusion}
\label{sec6}

In this study, we investigate the restless contextual MP-MAB for allocating channels to SUs in an OSA scenario.
We take channel state information as context because it can be used to characterize channel noise.
A linear model and a neural network are devised to model the correlation between context and channel rewards' perturbation that is considered as the effect of channel noise.
The estimated perturbation is employed to adjust the channel's upper confidence bound to derive noise-aware policies.
Numerical experiments on simulated OSA scenarios demonstrate that the proposed policies outperform context-free bandits in terms of regret and sub-optimal channel selections.

In future work, we wish to extend our policies to more complex contextual settings.
This involves improving the current neural network for better efficiency and scalability.
In addition, the current MP-MAB works on a centralized BS.
How to adjust it for a decentralized architecture is worth of further investigations.

\section*{Acknowledgment}
The work was supported by the National Natural Science Foundation of China (Grant No.:~61602356).

\bibliographystyle{IEEEtran}
\bibliography{myref}

\end{document}